\DocumentMetadata{
	lang		= eng-US, 	% default
	pdfstandard	= A-3u,		% A-2b, A-2u, A-3b, A-3u. Check that your figures also meet the standard you choose.
	pdfversion	= 1.7, 		% an older version of PDF, but very widely supported
}

\documentclass[colorlinks,upint,subscriptcorrection,varvw,hyphenate,balance,greek,russian,vietnamese,german]{asmeconf} % pdftex

\usepackage{float}
\usepackage{amsfonts}

\usepackage{array}
\hypersetup{%
	pdfauthor={John H. Lienhard},									  % <=== change to YOUR name
	pdftitle={ASME Conference Paper LaTeX Template},                  % <=== change to YOUR pdf file title
	pdfkeywords={ASME conference paper, LaTeX template, BibTeX style},% <=== change to YOUR pdf keywords
	pdfsubject = {Describes the asmeconf LaTeX template},			  % <=== change to YOUR subject
}
\allowdisplaybreaks % from amsmath package, allows multiline equations to break across pages (delete if not wanted)
\begin{document}

% Change these fields to the right content for your conference.
% You can comment these out if for some reason you don't want a header.
% Use title case for the conference name (first letters capitalized), not all capitals

\ConfName{Proceedings of the ASME 2026\linebreak International Mechanical Engineering Congress and Exposition}
\ConfAcronym{IMECE2026}
\ConfDate{September 03--05, 2026} 
\ConfCity{Chennai, India}
\PaperNo{IMECE2026-188997}

% Units of measure (e.g., cm) and other specialty lowercase terms in the title should be 
%   enclosed in \NoCaseChange{...} to maintain lower case type.
%   The rest of the title will automatically be set in all capital letters.
%
%	\title{Place Title Here: Place Subtitle After Colon} 

\title{Guidance and Navigation Control of Unmanned Surface Vehicle } % <=== replace with YOUR title
 
%   Put author names into the order you want. Use the same order for affiliations.
%   \affil{#} tags the author's affiliation to the address in \SetAffiliation{#}.
%   No space between last name and \affil{#}, separate names with commas.
%
%	For a sole author or a single affiliation for all authors, {#} may be left empty, i.e. \affil{} and \SetAffiliation{} (but not with [grid] option!)
%
%   \CorrespondingAuthor{email} follows that author's affiliation, no spaces before or after 
%   If multiple corresponding authors, put both email addresses in the same command and place after both authors.
%
%   \JointFirstAuthor, if applicable, follows the affiliation of the relevant authors, no spaces.

\SetAuthors{
    Ninad D. Joshi\affil{1}, 
    Ashisranjan Pradhan\affil{1},
	Chandrakiran Nakka\affil{1}, 
	Himabindu Allaka\affil{1},
    Ashok Kumar Pandey\affil{1}
	}

\SetAffiliation{1}{Department of Mechanical and Aerospace Engineering, IIT Hyderabad}

\maketitle

%%% Use this footnote for tracking various versions of your draft. Change text to suit your own needs. 
% %%% \date{..} calls the same command. 
% \versionfootnote{Documentation for \texttt{asmeconf.cls}: Version~\versionno, \today.}% <=== Delete before final submission.

%%% Change the following to your keywords.  Keywords are automatically printed at the end of the abstract.
%%% This command MUST COME BEFORE the end of the abstract.
%%% If you don't want keywords, leave the argument of \keywords{} empty (or use the abstract* environment)


%%%%%  End of fields to be completed. Now write your paper. %%%%%%%%%%%%%%%%%%%%%%%%%%%%%%%%%%%%%%%%%%%

%%%%%  ABSTRACT  %%%%%%%%%%%%%%%%%%%%%%%%%%%%%%%%%%%%%%%%%%%%%%%%%%%
%%
%% Abstract should be 200 words or less
\begin{abstract}
This study presents the modeling, control design, and simulation of a simplified 3-DOF Unmanned Surface Vehicle (USV) for guided autonomous waypoint-based navigation in the horizontal plane. A maneuvering model consisting of surge, sway, and yaw dynamics with linear damping and simplified disturbance representation is developed. A complete control framework is implemented in Python, including PID surge and heading control with Line of Sight (LOS) guidance for trajectory tracking. In addition, a Neural Network (NN)-based control approximation trained on PID-generated data and a Hybrid Fuzzy–PID controller are implemented. Thruster allocation with actuator constraints is considered for realistic operation. Simulation results demonstrate satisfactory waypoint tracking with low steady-state error and limited cross-track deviation under the given conditions. A comparative analysis shows that the NN-based and Fuzzy–PID approaches show smoother control response within the simulation framework, indicating their potential for preliminary evaluation in autonomous marine navigation.

\end{abstract}

%%%%%%%%%  NOMENCLATURE (OPTIONAL) %%%%%%%%%%%%%%%%%%%%%%%%%%%%%%%%%
%%
%% To change space between the symbols and  definitions, use \begin{nomenclature}[Xcm] where X is a number 
%% The unit cm can be replaced by any LaTeX unit of dimension: pt, in, ex, em, pc, etc.
%% Default is 2em.
%%
%% \EntryHeading{..} produces an italicized subheading in the nomenclature list, e.g., \EntryHeading{Greek letters}

\begin{nomenclature}
\EntryHeading{Roman letters}
\entry{$u$, $v$, $r$}{Surge, Sway [m s$^{-1}$]}
\entry{$x$,$y$, $z$}{Global Positions [Meters ]}
\entry{$\psi$}{Heading Angle [Degrees]}
\entry{$M$}{Mass+Added Mass Matrix[6$\times$6]}
\entry{$C(\nu)$}{Coriolis Matrix[6$\times$6]}
\entry{$D(\nu)$}{Hydrodynamic Damping Matrix[6$\times$6]}
\entry{$\tau$}{Control Forces and Moments Vector}
\entry{$\tau_{d}$}{Disturbances Vector}

% \entry{$\vec{q}$}{Heat flux vector [W m$^{-2}$]}
\entry{$\phi, \theta$}{Roll and Pitch Angles [rad]}
\entry{$\psi_d$}{Desired Heading Angle [rad]}

\entry{$w$}{Heave Velocity [m s$^{-1}$]}
\entry{$p, q$}{Roll, Yaw Rate and Pitch Rates [rad s$^{-1}$]}

\entry{$\eta$}{Position and Orientation Vector}
\entry{$\nu$}{Velocity Vector}

\entry{$e_{\psi}$}{Heading Error   [rad]}
\entry{$e_{ct}$}{Cross-Track Error  [m]}

\entry{$d_{wp}, d_x, d_y$}{Distance and Relative Position to Waypoint [m]}

\entry{$P_i$}{Waypoint Coordinates (i to n)}

\entry{$\tau_{surge}$}{Surge Force [N]}
\entry{$\tau_{yaw}$}{Yaw Moment [Nm]}

\entry{$X, Y, \hat{Y}$}{Input, Output, and Predicted Vectors}

\entry{$W^{(l)}, b^{(l)}$}{Weights and Biases of Layer $l$}

\entry{$L, N$}{Loss Function and Number of Samples}
\entry{$e, \dot{e}$}{error, change error}
\entry{$K_p, K_i, K_d$}{PID Gains}
\entry{$K_{\psi}, K_r$}{Heading and Yaw Damping Gains}

\entry{$\tau_{NN}$}{Neural Network Control Output}

\entry{$d$}{Gaussian Disturbance}
\entry{$t$}{Simulation Time [s]}

% \EntryHeading{Greek letters}
% \entry{$\alpha$}{Thermal diffusivity [m$^2$ s$^{-1}$]}
% \entry{$\nu$}{Kinematic viscosity [m$^2$ s$^{-1}$]}

% \EntryHeading{Dimensionless groups}
% \entry{Pr}{Prandtl number, $\nu/\alpha$}
% \entry{Sc}{Schmidt number, $\nu/\mathcal{D}_{1,2}$}

% \EntryHeading{Superscripts and subscripts}
% \entry{b}{bulk value}
% \entry{$\infty$}{free stream value}
\end{nomenclature}

%%%%%%%%%  BODY OF PAPER %%%%%%%%%%%%%%%%%%%%%%%%%%%%%%%%%

\section{Introduction}

Autonomous marine vehicles have gained significant attention in recent years due to their wide range of applications, including environmental monitoring, weather observation, seabed surveys, and various defense applications. Among these, accurate trajectory tracking is essential for ensuring the safe and efficient operation of unmanned surface vehicles (USVs). In this context, the present study proposes a comprehensive framework for the design and implementation of multiple control strategies, including Proportional-Integral-Derivative (PID), Neural Network (NN, and Hybrid Fuzzy–PID controllers.

For USVs, when the Froude number exceeds approximately 0.4, wave-making effects introduce significant nonlinearities in the hydrodynamic response. This necessitates the design of robust and reliable feedback control systems. However, due to the inherent underactuation of USVs, which are not equipped to control all six DOF, reduced-order models are commonly adopted. In this work, the system is therefore modeled using longitudinal three-DOF dynamics, and the control algorithm is formulated based on a maneuvering model \cite{Fossen2011}, \cite{Fossen1998}.

Fan et al. \cite{Yunsheng2015}. proposed a fuzzy self-adaptive PID controller for USV course control to improve heading regulation under nonlinear and disturbance-prone marine conditions. The study first developed a 3-DOF planar motion model of the USV and validated it using turning test data. A fuzzy inference mechanism was then used to tune the PID gains online based on the heading error and its rate of change. Their simulation results showed that the proposed controller achieved faster settling, lower overshoot, and better disturbance rejection than a conventional PID controller, particularly in the presence of wind, wave, and current disturbances. This work highlights the practical advantage of combining the simplicity of PID with the adaptability of fuzzy logic for marine vehicle heading control.

Li et al \cite{Lin2024} proposed a BP neural network–based self-tuning PID controller for multi-USV leader–follower formation control. The method uses a dual-channel structure to adapt the PID gains online for speed and heading regulation, allowing follower vessels to maintain the desired relative distance and angle with respect to the leader. Simulation results showed that the proposed controller could successfully achieve and maintain the desired formation without manual retuning of PID gains. However, the study also noted that although the BPNN-PID approach reduces tuning effort, it requires longer training time and may show weaker formation stability compared to a well-tuned conventional PID controller. Overall, the work demonstrates the potential of intelligent adaptive tuning for cooperative USV control, while also highlighting the need for further robustness improvement. 

Zhan et al.\cite{Zhan2025} (2025) proposed a robust heading control framework for unmanned surface vessels based on a Variable-Universe Fuzzy PID controller integrated with an improved Smith Predictor. The variable-universe mechanism dynamically adjusts the fuzzy domain according to the tracking error and error rate, improving adaptability and control precision under nonlinear and time varying conditions. A hybrid BAS-HSA-GA optimization algorithm was employed to tune controller parameters automatically. Simulation and experimental results demonstrated enhanced disturbance rejection, reduced steady-state error, and improved tracking performance compared with conventional PID and existing fuzzy control methods.

Kianejad et al. \cite{Kianejad2019} (2019) investigated the prediction of ship roll added-mass moment of inertia using Computational Fluid Dynamics (CFD) simulations based on a Harmonic Excited Roll Motion (HERM) technique. The study examined the influence of excitation frequency, forward speed, roll angle, appendages, and scale effects on hydrodynamic coefficients. Results showed that the non-dimensional added-mass moment of inertia varied significantly with operating conditions and could range between 10 and 32 percent of the ship's roll mass moment of inertia. The authors demonstrated that accurate estimation of hydrodynamic parameters improves the reliability of ship motion prediction compared with conventional constant-coefficient assumptions. Their work highlights the importance of incorporating realistic hydrodynamic effects into marine vehicle dynamic models for improved simulation fidelity and future control-system development.

We design the control system progressively, starting with classical controllers and then advancing toward more optimal strategies such as Neural Networks and Hybrid Fuzzy-PID control. First, PID controllers are implemented as the baseline approach. A surge-speed PID controller regulates the forward velocity of the USV, ensuring stable propulsion under varying disturbances. Similarly, a heading PID controller regulates the yaw angle $\psi$, with appropriate wrapping of the heading angle to ensure smooth directional transitions between $-\pi$ and $\pi$.

In this study, a simplified USV model is considered to establish the control framework and evaluate the preliminary trajectory-tracking behavior. The present formulation serves as an initial validation step before incorporating more realistic physical and hydrodynamic parameters.

Building upon this, a Neural Network (NN) model is trained on error data obtained from the PID controller, and an NN-based control system is developed using this model. Simultaneously, the PID gains ($K_p$, $K_i$, $K_d$) are adjusted according to fuzzy rules, and using this approach a hybrid control model is implemented.

To evaluate and compare the performance and reliability of the various control algorithms, the time histories of the surge, sway and yaw responses were analyzed. These responses provide insights into tracking accuracy, transient behavior, and stability characteristics under different control strategies.

\section{Mathematical Model}
The mathematical model used in this study is based on the formulation proposed by Thor I. Fossen \cite{Fossen2011}, and is described as follows: 

\subsection{Kinematic Model}

The position and orientation of the USV in the earth-fixed frame are represented by the vector
\begin{equation}
\boldsymbol{\eta} =
\begin{bmatrix}
x & y & \psi
\end{bmatrix}^{\top}
\end{equation}

where $x$ and $y$ denote the position of the vessel and $\psi$ represents the yaw angle.

The body-fixed velocity vector is defined as

\begin{equation}
\boldsymbol{\nu} =
\begin{bmatrix}
u & v & r
\end{bmatrix}^{\top}
\end{equation}

where $u$, $v$, and $r$ denote surge velocity, sway velocity, and yaw rate respectively.

The kinematic relationship between the earth-fixed and body-fixed frames is given by
\begin{equation}
    \dot{\boldsymbol{\eta}} = R(\psi)\,\boldsymbol{\nu}
    \label{eq:kinematics}
\end{equation}
where $R(\psi)$ is the rotation matrix defined as

\begin{equation}
R(\psi)=
\begin{bmatrix}
\cos\psi & -\sin\psi & 0\\
\sin\psi & \cos\psi & 0\\
0 & 0 & 1
\end{bmatrix}    
\end{equation}

The rotation matrix converts the body-fixed velocities $(u,v)$ into global-frame velocities $(\dot{x},\dot{y})$.
\subsection{Dynamic Model}

The dynamic behavior of the USV in the body-fixed frame can be expressed as
\begin{equation}
    M\,\dot{\boldsymbol{\nu}}
    + C(\boldsymbol{\nu})\,\boldsymbol{\nu}
    + D(\boldsymbol{\nu})\,\boldsymbol{\nu}
    + g(\boldsymbol{\eta})
    = \boldsymbol{\tau} + \boldsymbol{\tau}_{d}
    \label{eq:dynamics}
\end{equation}

For the present study, a simplified dynamic model of the USV is considered for preliminary evaluation and comparison of controller performance under basic motion conditions. The formulation is limited to three degrees of freedom, including surge, sway, and yaw, while higher-order motions are neglected. Only the rigid-body mass and inertia terms are retained in the system model, and added mass effects are neglected. The Coriolis and centripetal matrix $C(\boldsymbol{\nu})$, the restoring force vector $g(\boldsymbol{\eta})$ are not included for computational simplicity. A linear hydrodynamic damping term is retained to represent basic dissipative effects, while higher-order nonlinear hydrodynamic effects and actuator dynamics are not considered.Environmental disturbances are represented as additive stochastic inputs to emulate uncertainty in a simplified manner. Therefore, the adopted model is intended for initial controller validation rather than high-fidelity dynamic representation.

Below are the principal characteristics of the USV used in the study:
\begin{table}[H]
\centering
\caption{Principal characteristics of the USV}
\label{tab:usv_parameters}
\renewcommand{\arraystretch}{1.2}
\resizebox{\columnwidth}{!}{%
\begin{tabular}{|p{3.2cm}|p{1.8cm}|p{4.2cm}|}
\hline
\textbf{Parameter} & \textbf{Symbol} & \textbf{Value} \\ \hline

Mass & $m$ & $300~\mathrm{kg}$ \\ \hline

Yaw moment of inertia & $I_z$ & $200~\mathrm{kg\cdot m^2}$ \\ \hline

Linear damping matrix & $D$ & $\mathrm{diag}(50,\;80,\;100)$ \\ \hline

\end{tabular}%
}
\end{table}

\subsection{State-Space Representation}

For control design, the nonlinear system is linearized around a constant cruising speed $u_0$. The resulting reduced-order model can be expressed in state-space form as
\begin{equation}
    \dot{x} = A x + B u
    \label{eq:ss3}
\end{equation}

\begin{equation}
    y = C x
    \label{eq:ss4}
\end{equation}
The system state vector is defined as

\begin{equation}
    x = [\, u,\ v,\ r,\ x,\ y,\ \psi \,]^{\top}
\end{equation}

This state-space representation provides a convenient framework for controller design and system analysis.
\section{Controller Designs}
The objective of the control system is to ensure accurate trajectory tracking of the unmanned surface vehicle (USV). The controller regulates both the surge velocity and the heading angle of the vehicle to follow a predefined trajectory.
Two control loops are considered:
\begin{itemize}
\item Surge speed control
\item Heading control
\end{itemize}
Three control strategies are investigated in this work: a classical PID controller, and a Hybrid Fuzzy–PID controller, a Neural Network based controller.
% \begin{figure}[h!]
%     \centering
%     \includegraphics[width=0.5\textwidth]{control_system.png}
%     \caption{Control system block diagram}
%     \label{fig:control_system}
% \end{figure}
\subsection{PID}
A proportional–integral–derivative (PID) controller is used as the baseline control strategy. The control input is computed based on the tracking error between the desired reference signal and the measured output.
The tracking error is defined as
\begin{equation}
    e(t) = r(t) - y(t)
\end{equation}

where $r(t)$ is the desired reference trajectory and $y(t)$ is the measured output of the system.
The PID control law is given by
\begin{equation}
u(t) = K_p e(t) + K_i \int e(t) dt + K_d \frac{de(t)}{dt}
\label{eq:PID Eq}
\end{equation}

Two PID controllers are implemented in the system. The first PID controller regulates the surge velocity of the USV, while the second controller regulates the heading angle. The appropriate tuning of the gains $K_p$, $K_i$, and $K_d$ is performed to achieve stable and accurate trajectory tracking.

\subsection{Hybrid Fuzzy-PID}

In order to improve the robustness of the control system under disturbances and nonlinear dynamics, a hybrid Fuzzy–PID controller is implemented. In this approach, fuzzy logic is used to adaptively adjust the PID gains based on the system error and the rate of change in error.

The classical PID controller computes control input as \eqref{eq:PID Eq}

where $e(t)$ represents the tracking error. In the proposed hybrid controller, the PID gains $(K_p, K_i, K_d)$ are not fixed but are dynamically adjusted using a fuzzy inference system.

The fuzzy controller uses two inputs: $e$, $\dot{e}$

Both inputs are represented using five linguistic variables: Negative Big (NB), Negative Small (NS), Zero (ZO), Positive Small (PS), and Positive Big (PB). These membership functions allow the fuzzy system to interpret the magnitude and direction of the tracking error.

The fuzzy inference system produces three outputs corresponding to the PID gains i.e. $K_p$, $K_i$, $K_d$

A rule-based system is used to adjust the gains depending on the operating condition of the vehicle. For example, when the magnitude of error is large, the fuzzy controller increases the proportional gain to improve the response speed. When the error is small, the controller increases the integral gain to reduce steady-state error. Similarly, the derivative gain is adjusted on the basis of the rate of change of error to improve the system damping.

The tuned PID gains are then used to compute the control forces applied to the USV. In the implemented control structure, the controller generates two control inputs: the surge force and the yaw moment. These control inputs are then distributed to the individual thrusters through a simple thruster allocation scheme.

By adaptively adjusting the PID gains using fuzzy logic, the hybrid controller provides improved trajectory tracking performance compared to a fixed-gain PID controller.
\begin{figure}[h!]
    \centering
    \includegraphics[width=0.5\textwidth]{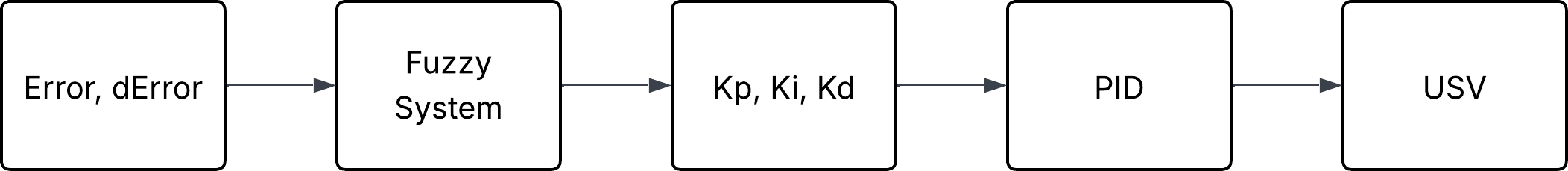}
    \caption{Flowchart for Fuzzy-PID Control}
    \label{fig:control_system}
\end{figure}
\subsection{Neural Network Controller}

A neural network based controller is developed to learn the nonlinear control law of the unmanned surface vehicle (USV). The neural network maps the system state and path-following errors directly to the required control forces for surge and yaw. 

\subsubsection*{Dataset Generation}
To train the neural network controller, a dataset was generated using simulations of the USV dynamic model combined with a PID control strategy. The purpose of this dataset is to capture the relationship between the vehicle state, path-following errors, and the corresponding control actions required to maintain trajectory tracking.

The dataset was obtained by running multiple simulations of the USV navigating through a sequence of predefined waypoints. During each simulation step, the system state and corresponding control commands were recorded.

\subsubsection*{State Variables Used in the Dataset}
At each simulation step, a feature vector representing the state of the system was constructed. The input feature vector consists of 18 variables describing both the vehicle motion and the path-following error.

\subsubsection*{Vehicle Pose and Orientation}
\begin{equation}
[x, y, z, \phi, \theta, \psi]
\end{equation}

\subsubsection*{Vehicle Velocity}

\begin{equation}
[u, v, w, p, q, r]
\end{equation}

\subsubsection*{Path Following Information}

\begin{equation}
[\psi_d, e_{\psi}, e_{ct}, d_{wp}, dx_{wp}, dy_{wp}]
\end{equation}

\subsubsection{Guidance Parameters}

\subsubsection*{Desired Heading}

Let the coordinates of two consecutive waypoints be

\begin{equation}
P_1 = (x_1, y_1), \qquad P_2 = (x_2, y_2)
\end{equation}

\begin{equation}
\psi_d = \tan^{-1}\left(\frac{y_2 - y_1}{x_2 - x_1}\right)
\end{equation}

\subsubsection*{Heading Error}

\begin{equation}
e_{\psi} = \psi_d - \psi
\end{equation}

To maintain numerical stability the error is normalized as

\begin{equation}
-\pi \leq e_{\psi} \leq \pi
\end{equation}

\subsubsection*{Cross-Track Error}

\begin{equation}
e_{ct} =
\frac{(x-x_1)(y_2-y_1) - (y-y_1)(x_2-x_1)}
{\sqrt{(x_2-x_1)^2 + (y_2-y_1)^2}}
\end{equation}

\subsubsection*{Distance to Waypoint}

\begin{equation}
d_{wp} =
\sqrt{(x-x_{wp})^2 + (y-y_{wp})^2}
\end{equation}

\subsubsection*{Relative Waypoint Distances}

\begin{equation}
d_x = x_{wp} - x, d_y = y_{wp} - y
\end{equation}

\subsubsection{Training Process}

The neural network was trained using supervised learning where the objective is to minimize the difference between predicted control outputs and recorded control actions.

\subsubsection*{Loss Function}

\begin{equation}
L = \frac{1}{N} \sum_{i=1}^{N} (y_i - \hat{y}_i)^2
\end{equation}

\subsubsection*{Optimization Algorithm}

The Adam optimizer was used to update network weights because of its adaptive learning rate properties.

\subsubsection*{Network Architecture}

The neural controller consists of fully connected layers:
\begin{figure}[H]
    \centering
    \includegraphics[width=0.5\textwidth]{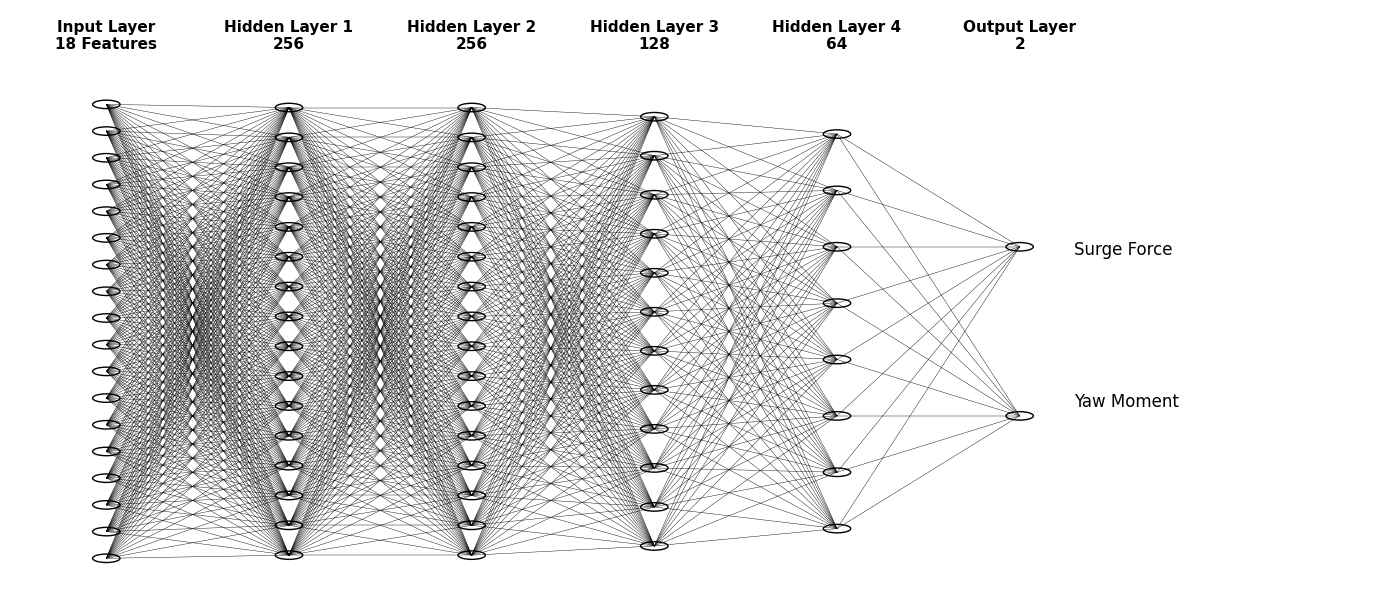}
    \caption{NN Architecture}
    \label{fig:nn_arch}
\end{figure}
% \begin{itemize}
% \item Input Layer : 18 neurons
% \item Hidden Layer 1 : 256 neurons
% \item Hidden Layer 2 : 256 neurons
% \item Hidden Layer 3 : 128 neurons
% \item Hidden Layer 4 : 64 neurons
% \item Output Layer : 2 neurons
% \end{itemize}

\subsubsection*{Neuron Computation}

\begin{equation}
y = w_{1}x_{1} + w_{2}x_{2} + \cdots + w_{18}x_{18} + b
\end{equation}

\subsubsection*{Activation Function}

\[
\text{ReLU}(x) = \max(0,x)
\]

\subsubsection*{Input Features}

\begin{equation}
X =
[x, y, z, \phi, \theta, \psi, u, v, w, p, q, r, \psi_d, e_{\psi}, e_{ct}, d_{wp}, dx_{wp}, dy_{wp}]
\end{equation}

\begin{equation}
X \in \mathbb{R}^{18}
\end{equation}

\subsubsection*{Neural Network Outputs}

\begin{equation}
Y =
\begin{bmatrix}
\tau_{surge} \\
\tau_{yaw}
\end{bmatrix}
\end{equation}

\subsubsection{Hybrid Control Strategy}

To improve stability, the neural network output is combined with a feedback controller.

\subsubsection*{Surge Adjustment}

\begin{equation}
\tau_{surge} = \tau_{NN}\cos(e_{\psi})
\end{equation}

\subsubsection*{Yaw Control Law}

\begin{equation}
\tau_{yaw} = \tau_{NN} + K_{\psi} e_{\psi} - K_{r} r
\end{equation}

\subsubsection{Control Signal Limits}

\subsubsection*{Surge Force}

\begin{equation}
20 \leq \tau_{surge} \leq 200
\end{equation}

\subsubsection*{Yaw Moment}

\begin{equation}
-60 \leq \tau_{yaw} \leq 60
\end{equation}

\subsubsection{Simulation Environment}

Disturbances are modeled as

\begin{equation}
d \sim \mathcal{N}(0,0.2)
\end{equation}
\subsubsection{Simulation Parameters}
\begin{align*}
\text{Simulation steps} &= 2000 \\
\Delta t &= 0.1 \ \text{s} \\
d &\sim \mathcal{N}(0, 0.2)
\end{align*}
% \begin{center}
% \[
% \begin{array}{rcl}
% \text{Simulation steps} & = & 2000 \\
% \Delta t & = & 0.1 \ \text{s} \\
% d & \sim & \mathcal{N}(0, 0.2)
% \end{array}
% \]
% \end{center}
% \begin{table}[h]
% \centering
% \begin{tabular}{|c|c|}
% \hline
% Parameter & Value \\
% \hline
% Simulation steps & 2000 \\
% Time step & 0.1 s \\
% Disturbance & Gaussian noise \\
% \hline
% \end{tabular}
% \end{table}

\section{Simulation Results and Discussion}
The performance of the proposed control strategies was evaluated through numerical simulations of the 3-DOF USV model. Three controllers were implemented and compared: a classical PID controller, a Hybrid Fuzzy–PID controller, and a Neural Network based controller. The objective of the controllers was to ensure accurate waypoint tracking of the USV while maintaining stable surge velocity and heading.
\begin{figure}[H]
    \centering
    \includegraphics[width=0.45\textwidth]{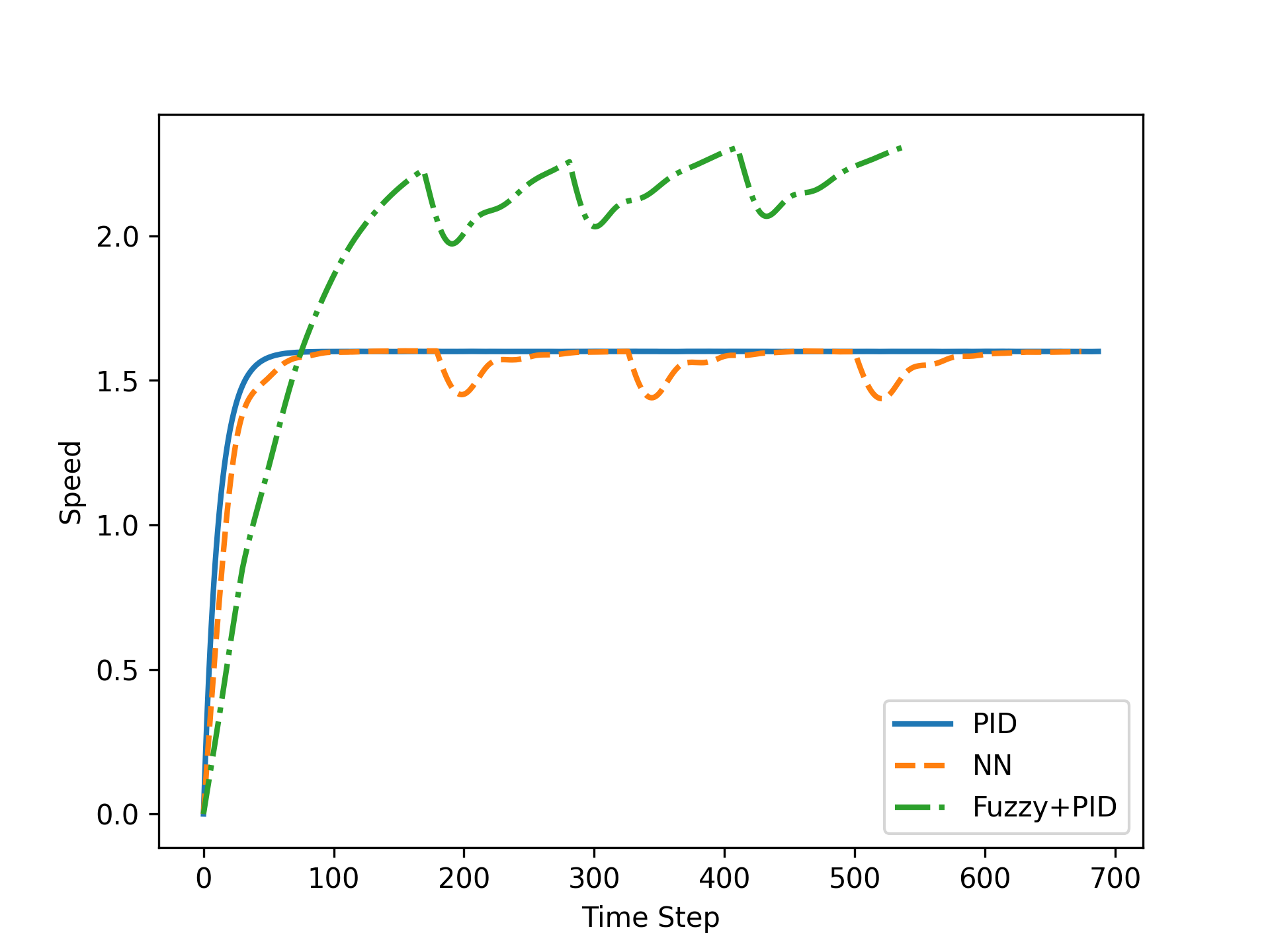}
    \caption{Surge speed control}
    \label{fig:surge_speed}
\end{figure}
Figure~\ref{fig:surge_speed} shows the surge speed generated by the three controllers during the simulation. The classical PID controller produces a large initial control spike, indicating aggressive transient behavior and higher actuator demand. Although the response stabilizes, such peaks may reduce actuator efficiency. The Hybrid Fuzzy–PID controller moderates the control effort by adapting gains online, but exhibits mild oscillations due to continuous gain adjustment and speed given by the hybrid fuzzy is not good as other two which consists somewhat larger error margin. The Neural Network controller generates the smoothest control response, with reduced peak magnitude and fewer oscillations. This indicates improved control smoothness and potentially lower actuator stress.
\begin{figure}[H]
    \includegraphics[width=0.45\textwidth]{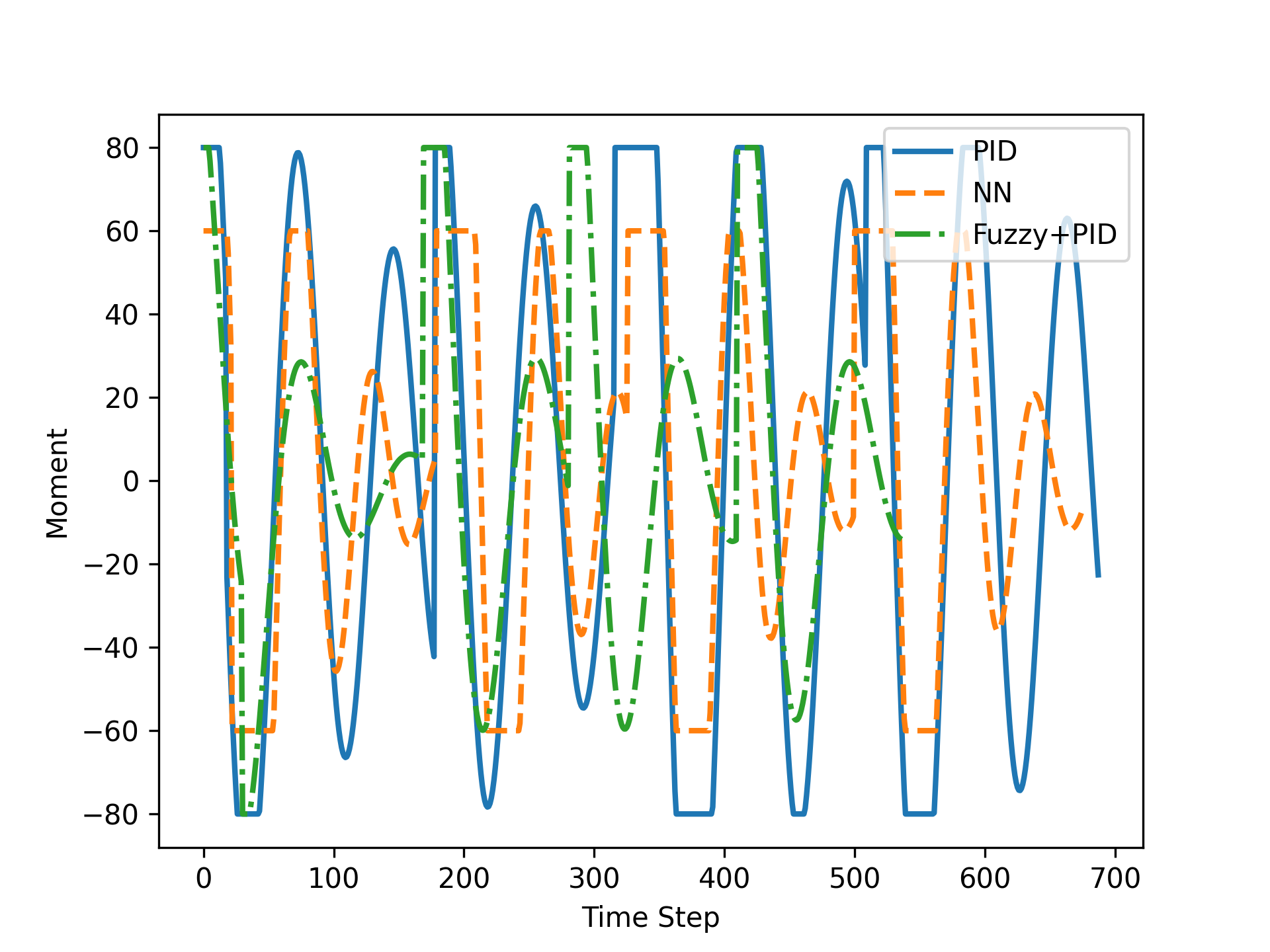}
    \caption{Yaw moment control}
    \label{fig:yaw_moment}
\end{figure}
Figure~\ref{fig:yaw_moment} illustrates the yaw control moment produced by the three controllers. The classical PID controller shows significant oscillations, reflecting aggressive heading corrections. The Hybrid Fuzzy–PID controller reduces the oscillation amplitude through adaptive gain tuning, resulting in a more stable response. The Neural Network controller exhibits the smoothest yaw moment profile with minimal oscillatory behavior, indicating improved heading stability and smoother actuation. The yaw moment is expressed in Newton-meters (Nm).
\begin{figure}[H]
    \centering
    \includegraphics[width=0.45\textwidth]{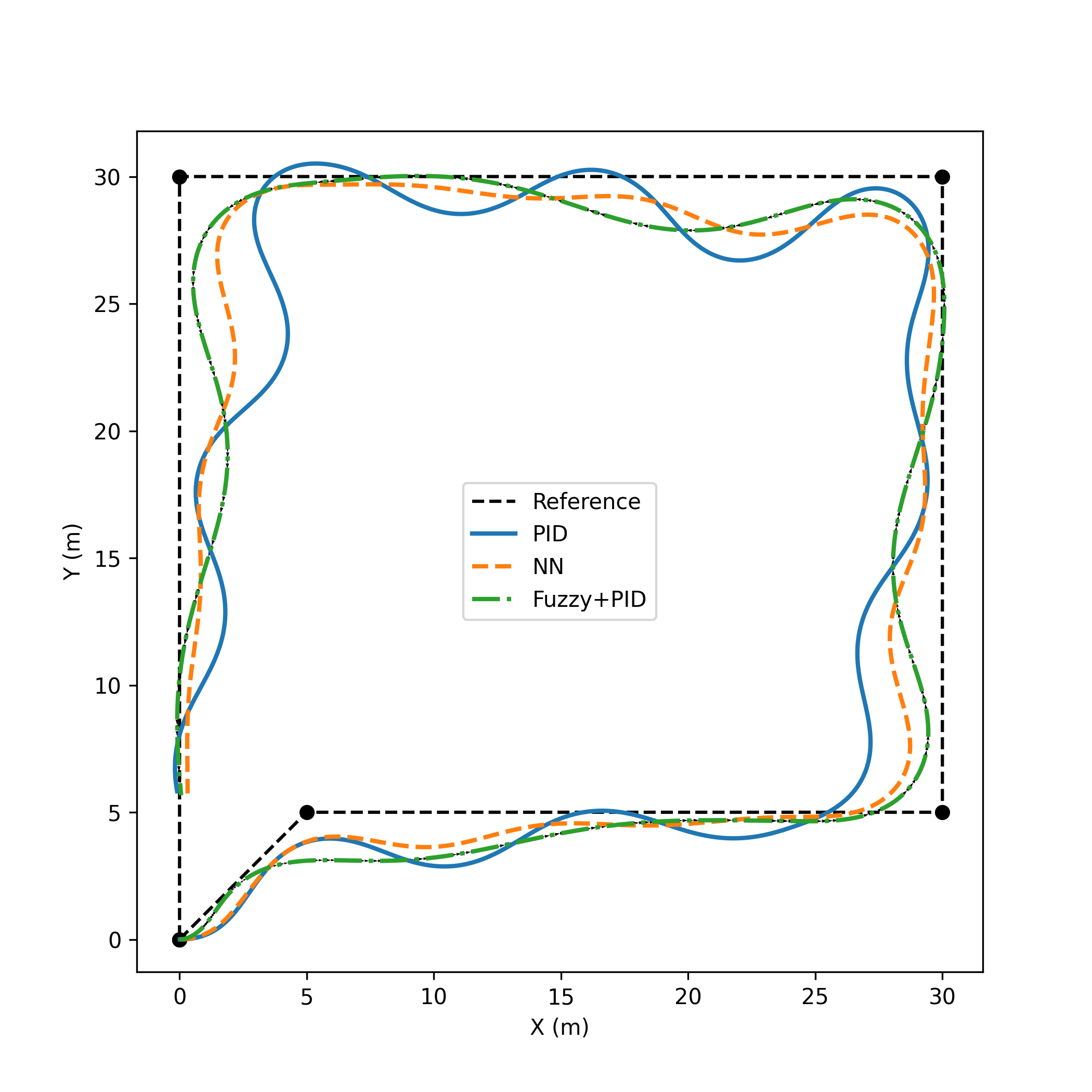}
    \caption{Comparison of trajectory path for various controllers}

        \label{fig:trajectory}
\end{figure}
Figure~\ref{fig:trajectory} presents the trajectory tracking performance of the USV for the three controllers. However, the classical PID controller exhibits noticeable deviations, particularly near waypoint transitions due to delayed heading correction. The Hybrid Fuzzy–PID controller improves tracking accuracy by adaptively tuning the gains, resulting in smoother path following. The Neural Network controller demonstrates the smoothest trajectory with reduced deviation from the reference path, primarily due to its ability to approximate the control behavior learned from PID-generated data. This results in improved path smoothness within the simulation framework.
\begin{figure}[H]
    \centering
    \includegraphics[width=0.45\textwidth]{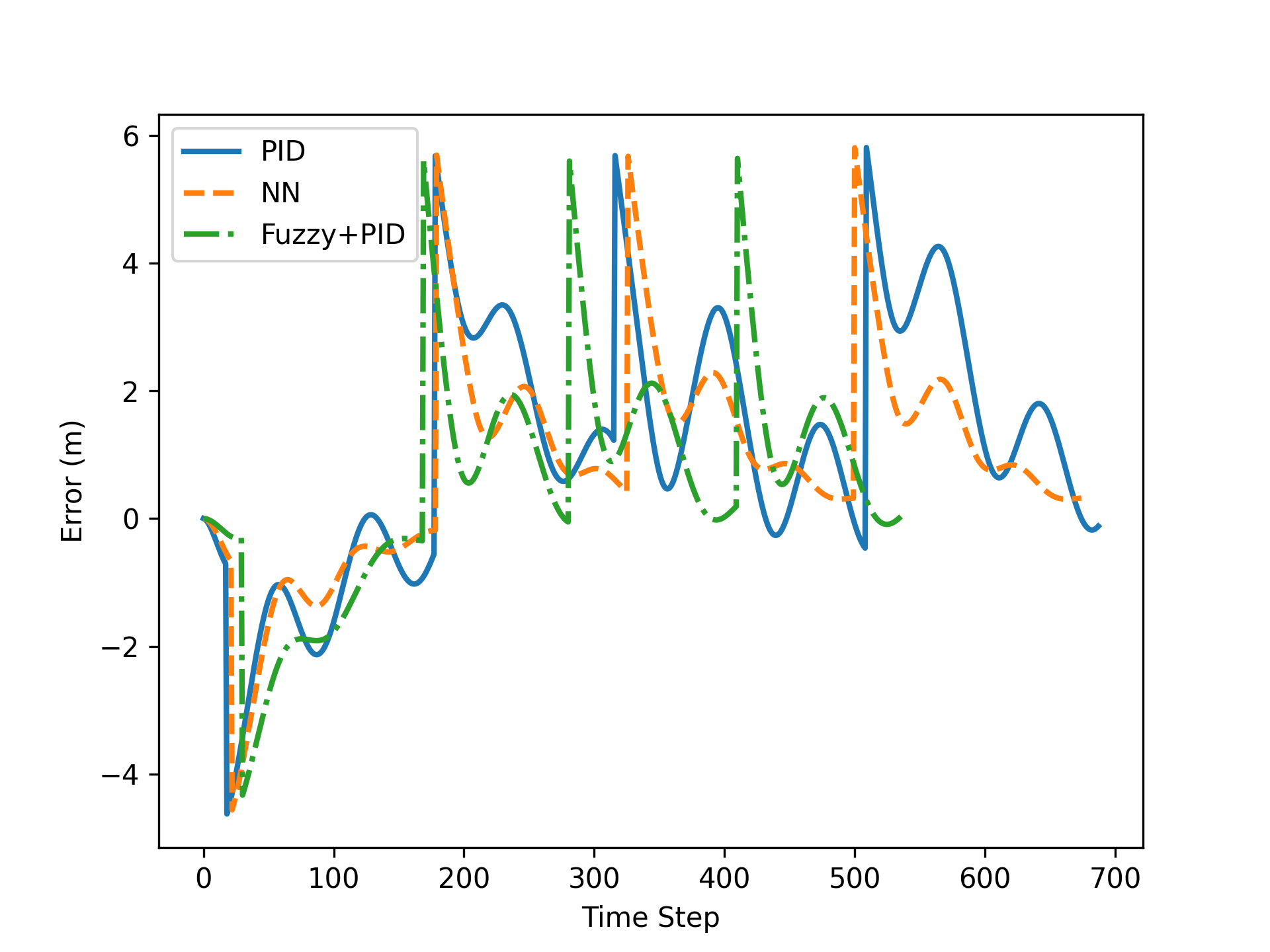}
    \caption{Cross-track error comparison}
    \label{fig:cross_track}
\end{figure}
Figure~\ref{fig:cross_track} shows the cross-track error during waypoint tracking for the three controllers. The classical PID controller exhibits larger deviations from the desired path, especially during transitions between waypoints. The Hybrid Fuzzy–PID controller reduces the magnitude of these deviations by adapting control gains based on the tracking error. The Neural Network controller achieves the lowest cross-track error with smoother variation, indicating improved lateral tracking performance. Peak errors are observed near waypoint transitions due to discontinuities in the desired heading.
\begin{figure}[H]
    \centering
    \includegraphics[width=0.45\textwidth]{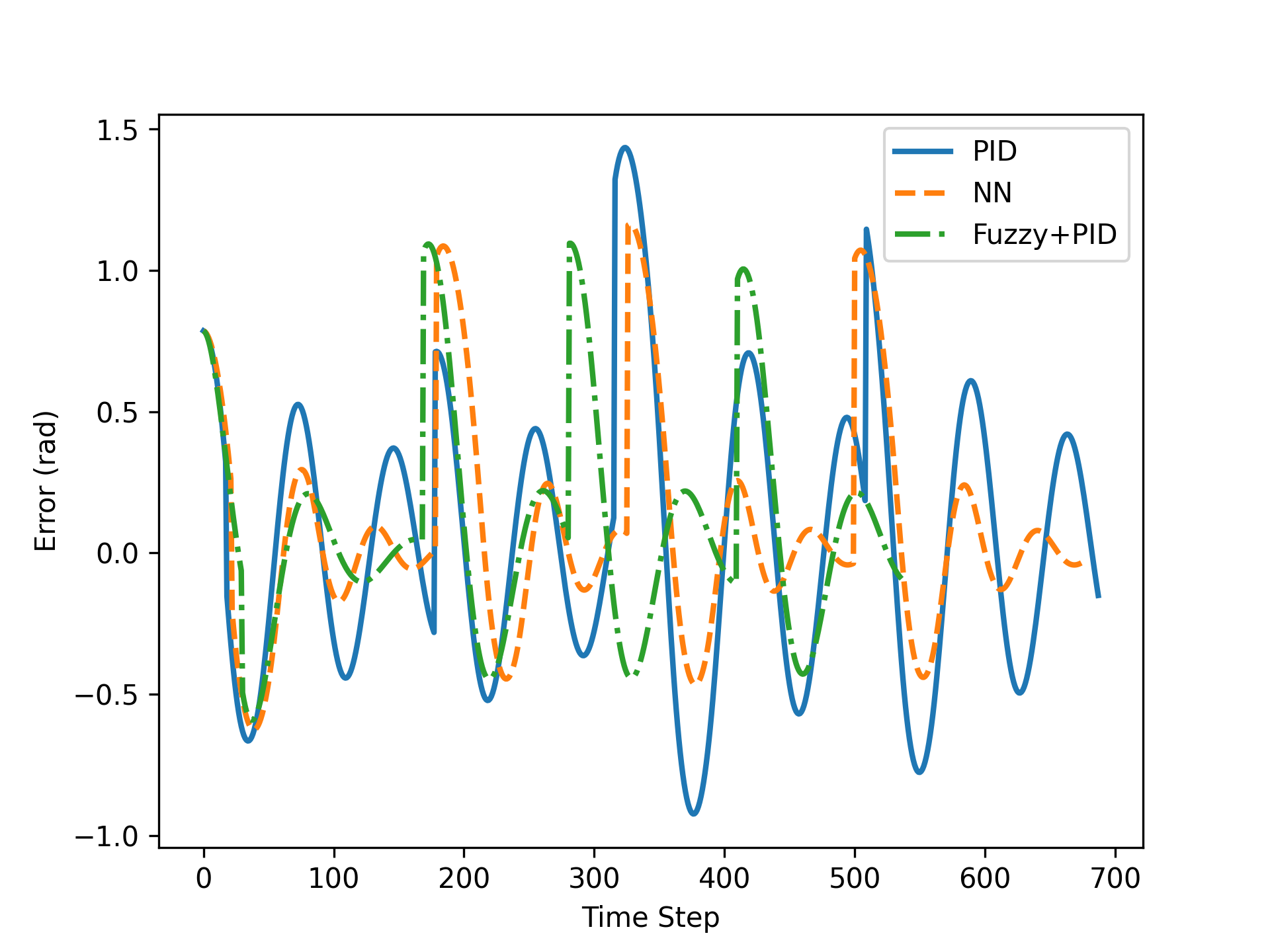}
    \caption{Heading error}
    \label{fig:heading_error}
\end{figure}
Figure~\ref{fig:heading_error} presents the heading error evolution during the simulation. The classical PID controller shows noticeable oscillations, corresponding to aggressive yaw corrections. The Hybrid Fuzzy–PID controller reduces these oscillations through adaptive gain tuning, resulting in improved damping characteristics. The Neural Network controller achieves the smallest heading error with smoother transitions, indicating better orientation control. The oscillatory behavior observed in the PID response corresponds to the variations in yaw moment shown in Figure 4.

\section{Conclusion}

This paper presented a comparative study of different control strategies for trajectory tracking of a 3-DOF unmanned surface vehicle (USV). A classical PID controller, a Hybrid Fuzzy–PID controller, and a Neural Network based controller were implemented and evaluated through numerical simulations. The controllers were tested for waypoint tracking, and their performance was analyzed using control force and moment, trajectory tracking, cross-track error and heading error under environmental disturbances are modeled as stochastic perturbations for preliminary robustness analysis.

Simulation results show that the Hybrid Fuzzy–PID controller and NN controller shows smoother response characteristics within the trained operating conditions compared to the conventional PID controller by adaptively tuning the control gains. NN controller demonstrates smoother response characteristics. These results highlight the potential of learning-based controllers for improving the robustness and accuracy of autonomous marine vehicle navigation.

The results in this study are limited to a simplified dynamic model and serve as a preliminary validation of the proposed control strategies.

\section{Future Work}
The present study is based on a simplified USV model intended for preliminary validation of the proposed control strategy. As part of future work, the model will be refined by incorporating realistic physical dimensions, mass–inertia properties, hydrodynamic coefficients, and actuator characteristics of the actual USV platform. This higher-fidelity representation is expected to provide more realistic simulation behavior and enable a more rigorous assessment of trajectory-tracking performance. 

However, since the neural network is trained on PID-generated data and evaluated under similar conditions, the observed improvements are limited to the scope of the current simulation framework. Future work will focus on incorporating more diverse training scenarios and exploring physics-informed learning approaches to improve generalization.

\section*{Acknowledgment}
The authors would like to express their sincere gratitude to their faculty for valuable guidance and suggestions throughout the course of this work. The support provided by the Department of Mechanical and Aerospace Engineering, IIT Hyderabad, is also greatly appreciated.

\bibliographystyle{asmeconf}
\bibliography{references}
\appendix
\end{document}